\documentclass{tADR2e}

\usepackage{graphicx}

\usepackage{bm}
\usepackage{color}
\usepackage{epsfig} % for postscript graphics files
\usepackage{mathptmx} % assumes new font selection scheme installed
\usepackage{times} % assumes new font selection scheme installed
\usepackage{amsmath} % assumes amsmath package installed
\usepackage{amssymb}  % assumes amsmath package installed
\usepackage{algorithm}
\usepackage{algorithmic}

\let\em\relax
\DeclareTextFontCommand{\em}{\it}
\let\emph\relax
\DeclareTextFontCommand{\emph}{\it}

\begin{document}

\jvol{xx}
\jnum{xx}
\jyear{20xx}
\jmonth{Month}

\articletype{Survey Paper}

\counterwithout{footnote}{page}

\title{
World Models and Predictive Coding for Cognitive and Developmental Robotics:\\
Frontiers and Challenges
}

\author{
    Tadahiro Taniguchi$^{a}$$^{\ast}$\thanks{$^\ast$Corresponding author. Email: taniguchi@ci.ritsumei.ac.jp\vspace{6pt}},
    Shingo Murata$^b$,
    Masahiro Suzuki$^c$,    
    Dimitri Ognibene$^{d,e}$,
    Pablo Lanillos$^{f,g}$,
    Emre Ugur$^h$,
    Lorenzo Jamone$^i$,
    Tomoaki Nakamura$^j$,
    Alejandra Ciria$^k$,
    Bruno Lara$^l$, and
    Giovanni Pezzulo$^m$\\
    \vspace{6pt}
    $^{a}${\em{Department of Information Science and Engineering, Ritsumeikan University,\\
    1-1-1 Noji-Higashi, Kusatsu, Shiga, Japan}}\\
    $^{b}${\em{Keio University, Japan}}\\
    $^{c}${\em{The University of Tokyo, Japan}}\\
    $^{d}${\em{Università Milano-Bicocca, Italy}}\\
    $^{e}${\em{University of Essex, UK}}\\
    $^{f}${\em{Donders Institute for Brain, Cognition and Behaviour, Netherlands}}\\
    $^{g}${\em{Cajal International Neuroscience Center, Spanish National Research Council, Spain.}}\\
    $^{h}${\em{Bogazici University, Turkey}}\\
    $^{i}${\em{Queen Mary University of London, UK}}\\
    $^{j}${\em{The University of Electro-Communications, Japan}}\\
    $^{k}${\em{National Autonomous University of Mexico, Mexico}}\\
    $^{l}${\em{Universidad Autónoma del Estado de Morelos, Mexico}}\\
    $^{m}${\em{Institute of Cognitive Sciences and Technologies, National Research Council of Italy, Italy}}\\ 
    \vspace{6pt}
    \received{v1.0 released April 2021}
}

\maketitle

\begin{abstract}
Creating autonomous robots that can actively explore the environment, acquire knowledge and learn skills continuously is the ultimate achievement envisioned in cognitive and developmental robotics. 
Importantly, if the aim is to create robots that can continuously develop through interactions with their environment, their learning processes should be based on interactions with their physical and social world in the manner of human learning and cognitive development.
Based on this context, in this paper, we focus on the two concepts of \textit{world models} and  \textit{predictive coding}.
Recently, world models have attracted renewed attention as a topic of considerable interest in artificial intelligence. 
Cognitive systems learn world models to better predict future sensory observations and optimize their policies, i.e., controllers. 
Alternatively, in neuroscience, predictive coding proposes that the brain continuously predicts its inputs and adapts to model its own dynamics and control behavior in its environment. Both ideas may be considered as underpinning the cognitive development of robots and humans capable of continual or lifelong learning. Although many studies have been conducted on predictive coding in cognitive robotics and neurorobotics, the relationship between world model-based approaches in AI and predictive coding in robotics has rarely been discussed. 
Therefore, in this paper, we clarify the definitions, relationships, and status of current research on these topics, as well as missing pieces of world models and predictive coding in conjunction with crucially related concepts such as the free-energy principle and active inference in the context of cognitive and developmental robotics. Furthermore, we outline the frontiers and challenges involved in world models and predictive coding toward the further integration of AI and robotics, as well as the creation of robots with real cognitive and developmental capabilities in the future.

    \medskip
    % Provide 3 to 6 keywords in small capitalization.
    \begin{keywords}
        World model; cognitive robotics; predictive coding; free-energy principle; active inference; deep generative models
    \end{keywords}
\end{abstract}

\section{Introduction}
\label{sec:introduction}
How can we develop robots that can autonomously explore the environment, acquire knowledge, and learn skills continuously?  Creating autonomous cognitive and developmental robots that can co-exist in our society has been considered an ultimate goal of cognitive and developmental robotics and artificial intelligence (AI) since the inception of these fields. Autonomous robots that can develop in the real world and collaborate with us may also be called embodied artificial general intelligence (AGI).
The recent success of artificial intelligence depends primarily on large-scale human-annotated data. However, human infants can acquire knowledge and skills from sensorimotor information through physical interactions with their environment and social interactions with others (e.g., their parents or caregivers). 
Importantly, the aim is to build robots that can continuously develop through embodied interactions, their learning process must be strongly based on their own sensorimotor experiences.
This autonomous learning process that occurs throughout development is also referred to as continual or lifelong learning \cite{lungarella2003developmental,lesort2020continual,DeLange2021continual}, and is considered the foundation for the emergence of both individual and social abilities necessary for robots with adaptive and collaborative capabilities.

Recently, {\it world models} have attracted renewed attention in artificial intelligence~\cite{friston2021world,ha2018world,hafner2020mastering,wu2022daydreamer}.
Now, the term ``world'' does not indicate the objective world but rather refers to a world understood from a robot's point of view\footnote{This viewpoint may be called a robot's subjective point of view of the world. Philosophically, however, whether robots can have a ``subjective'' point of view remains controversial~\cite{kiverstein2007could}. Therefore, we describe this point of view simply as ``a robot's point of view''.}. This idea corresponds to that of \textit{Umwelt} proposed by Uexk\"{u}ll~\cite{Uexkull}. Umwelt, literally around-world, meaning environment or surroundings, refers to the self-centered world of an organism perceived through its species-specific sensors\footnote{Importantly, the relationship between Umwelt and world modeling was suggested in semiotics. Sebeok pointed out that the closest equivalent of Umwelt in English is ``model''~\cite{sebeok2001biosemiotics}. An Umwelt is created and constructed through a functional cycle, which includes 1) anticipation of a perceptual cue, 2) perception, 3) working out a relation between the perception and action (either simply executing a habit or using representation, or modeling anew), and 4) action (operation)~\cite{kull2009umwelt}.}.
Therefore, notably, the world model is different from the bird's-eye model of the world that was aimed to build in good-old-fashioned AI and criticized later~\cite{Brooks1991}\footnote{Also, notably, the world-model approach is different from behavior-based robotics~\cite{arkin1998behavior}, which does not learn world models.}.
A cognitive system learns a world model to predict its future sensory observations better and optimize its policies, also referred to as controllers. Note that although typically the term ``world model'' is used to denote the spatiotemporal dynamics of the external environment, it could also equally apply to bodily dynamics (including interoceptive signals from inside the body) and the social environment.
This world-model view entails previous ideas and results, such as the effect of behavioral feedback on sensory sampling and perceptual learning \cite{verschure2003environmentally} and the resulting acquisition of self-centered, yet efficient, representations induced by an active perception strategy on the part of an agent \cite{ognibene2014ecological}.

\textit{Predictive coding} is another related theory that recently has become more and more influential \cite{clark2013a}. It is heavily influenced by Helmholtz's early theories of perception as a process driven by learning, knowledge, and inference \cite{helmholtz1867handbuch}.  Predictive coding proposes that the brain  infers the external causes of sensations by continuously predicting its input through top-down signals and adapts to minimize prediction error ~\cite{Rao1999,Friston2005}. This substantiates the idea that the brain might use an adaptive world model to support perception. The free energy principle (FEP) also proposes a similar vision. It argues that our brain supports both perception (perceptual inference) and action (active inference) using a form of variational Bayesian inference; in particular, using (variational) free energy, it assesses the quality of the prediction and its conformity to prior beliefs~\cite{Parr2022}. These ideas, which are currently influential in neuroscience and cognitive science, are also used in cognitive and developmental robotics, neurorobotics~\cite{ciria2021predictive,oliver2021empirical,lanillos2021active}, and artificial intelligence to develop neurodynamics realizing adaptive behaviors and social perception~\cite{tani2016exploring}.

%Although the world model-based approach is promising in cognitive and developmental robotics, the many applications and studies on world models tend to be limited to simulations, especially in AI. 
Although such a learning-driven world model-based approach is promising in cognitive and developmental robotics, the many applications and studies of world models tend to be limited to simulation studies or adopt an offline pretrained world model \cite{ibarz2021train}. 
Meanwhile, many studies based on predictive coding have been conducted in the field of cognitive robotics and neurorobotics. However, the relationship between the world model-based approach in AI and the predictive coding-based approach in robotics has rarely been discussed in an integrated manner. We believe that clarifying the definition, relationship, current state of the art, notable research gaps in work on  world models, predictive coding, free-energy principle, and active inference in the context of cognitive and developmental robotics is important for further progress in this field. Based on the current status, we elucidate the frontiers and challenges toward this holy grail in cognitive and developmental robotics.

In this survey paper, we aim to build bridges and clarify the challenges and frontiers of world models and predictive coding in cognitive robotics. 
The remainder of this paper is structured as follows. Section 2 provides a working definition of each key concept. Section 3 describes prior works related to the concepts and clarifies state of the art. Section 4 describes some notable challenges. 
Some additional discussion is provided in Section 5, and we conclude the work in Section 6.

\begin{figure}[tb]
\centering
\includegraphics[width=1.0\linewidth]{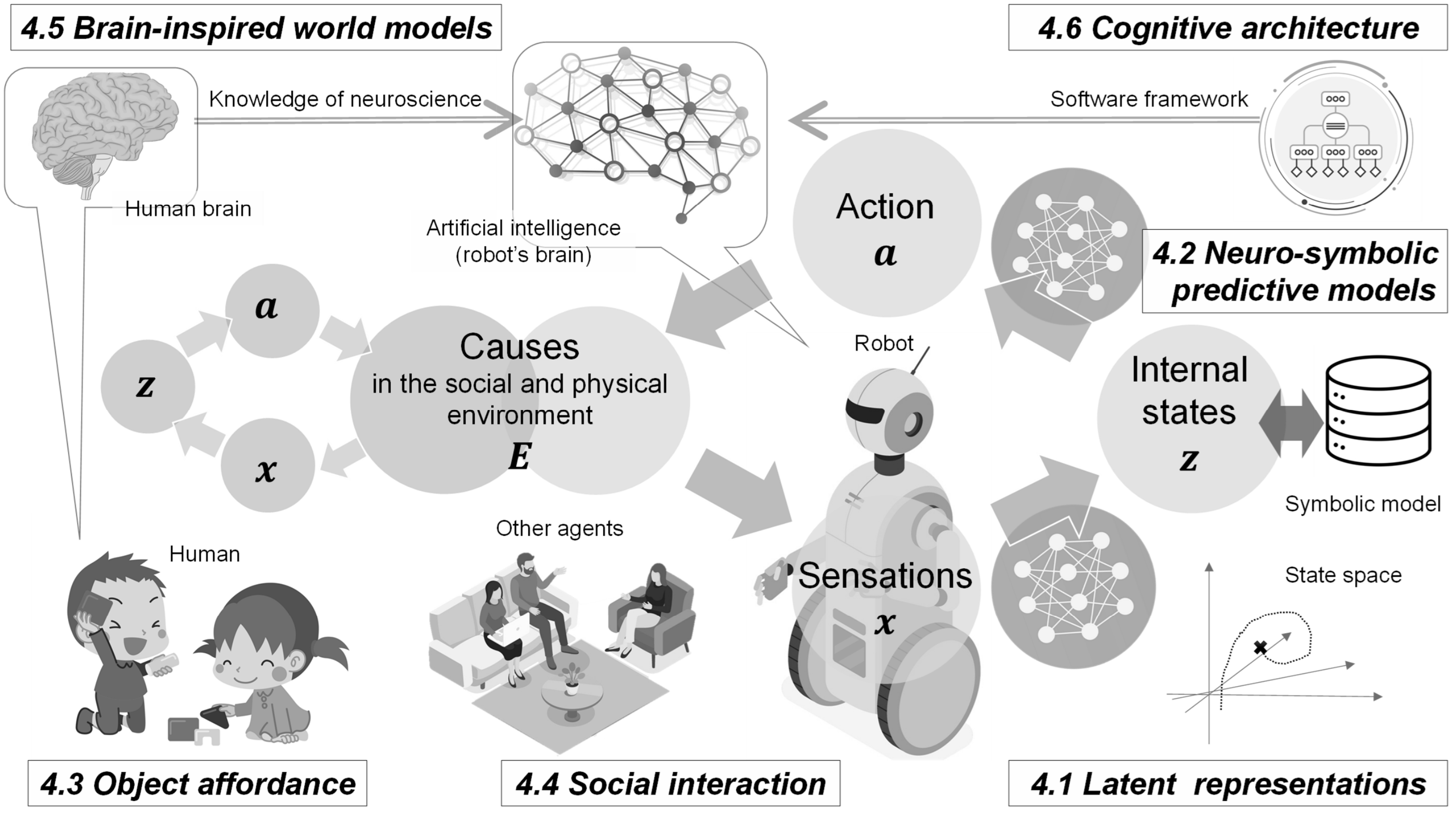}
\caption{Overview of challenges and relationships between topics described in this survey. A robot, similar to a human, receives \emph{sensations} $x$, infers \emph{internal states} $z$, exhibits \emph{actions} $a$, and affects \emph{causes} $E$ in the social and physical environment. The initial problem is determining the models and architecture that a robot uses to efficiently and effectively learn \emph{latent representations}. An approach to this problem uses a \emph{neuro-symbolic predictive model}, which combines neural network and symbolic models. The notion of \emph{object affordance} highlights the importance of object-centric representation learning and the coupling of action and perception. \emph{Social interaction} with other agents is also an important area of research. Developing artificial intelligence for cognitive and developmental autonomous robots based on knowledge of neuroscience, i.e.,  \emph{brain-inspiredworld models}, is promising. Creating \emph{cognitive architecture} and developing and sharing software frameworks for this purpose will also be an important frontier.}
\label{fig:overview}
\end{figure}

\section{Working definition}
\subsection{World model}
World models describe the {\it internal models} of an agent, which encodes how world states evolve, respond to agents' actions, and relate to a given sensory input~\cite{ha2018recurrent,friston2021world}. The term {\it world model} dates back to the beginnings of artificial intelligence and robotics ~\cite{nilsson1984shakey}. Early research in machine learning studied how an agent could independently acquire and adapt a world model to \cite{schmidhuber1990making, sutton1990integrated}. Currently, it usually refers to predictive models~\cite{hafner2019dream}, which are mainly encoded using deep neural networks.

In recent years, advancements in the studies on deep neural networks have enabled self-supervised (or unsupervised) learning\footnote{Self-supervised learning is a type of unsupervised learning that aims to accomplish a task by learning to predict or classify any part of unsupervised data from any other part. In contrast to self-supervised learning, unsupervised learning includes clustering.} of large-scale world models directly from observations (sensory inputs)~\cite{ha2018recurrent,levine2016end,van_der_Himst_2020}, and these models have been applied in various areas of artificial intelligence, including reinforcement learning (RL). 
World models allow agents to perform a sample-efficient prediction of the present state of the world and enable the prediction of future states, which further enables efficient planning (equivalent to model-based reinforcement learning or control). A compact internal representation further enables planning in an efficient low-dimensional space.

The key elements of world models are  \emph{prediction} and \emph{inference}\footnote{The use of these terms is sometimes incongruent in the literature on statistics and machine learning; the word {\it inference} is also used to describe prediction or substituted by other concepts, such as \emph{encoding} and \emph{decoding} in the literature on (variational) autoencoders.}. \emph{Prediction} is the probabilistic process of generating the \emph{observation} $x$ given the \emph{state} (or \emph{representation}) $z$, whereas \emph{inference} is the process of obtaining a state representation $z$ from an observation $x$ in a probabilistic manner. In real-world settings, observations $x$ are large (high-dimensional, e.g., images) and provide only partial information about the world (partial observability). At the same time, the latent representation $z$ is assumed to represent the internal state of the world.
In a static case, these can be summarized as the generative processes of probability distributions as follows.
\begin{align}
\makebox[8em][l]{Prediction:} &x\sim p(x|z) \nonumber\\
\makebox[8em][l]{Inference:} &z\sim q(z|x), \label{eq_gm}
\end{align}
where $p$ is a generative model and $q$ is an inference (or recognition) model\footnote{In classical formulations of control theory \citep{kalman1960filtering}, AI \citep{cassandra1994acting}, and probabilistic robotics \citep{thrun2005probabilistic}, the inference step is performed by exactly inverting the prediction probability $p$ using the Bayes theorem. However, this poses several computational challenges and is often intractable. For dealing with such complexity, sampling-based approximate inference can be performed using Monte Carlo methods~\cite{chen2003filtering,thrun2002particle}. In the context of world models and predictive coding, variational Bayes approaches are often preferred~\cite{Parr2022}. Variational Bayes involves the definition of an approximate inference distribution $\hat{q}$. Amortized inference allows us to approximate the $q(x)$ using a neural network, which is called an inference network, and obtain an inference model $q(x|a)$~\cite{kingma2013auto}.}. These models are considered parameterized in deep neural networks. When these models are trained simultaneously, generative approaches (e.g., variational autoencoders~\cite{kingma2013auto}) are employed~\cite{ha2018recurrent}. There are also cases where only an inference model is trained, in which case a discriminative approach (e.g., contrastive learning~\cite{laskin2020curl}) is used\footnote{However, recently, studies have also shown that contrastive learning can be interpreted as a generative approach~\cite{nakamura2022self}. Therefore, the boundary between generative and discriminative approaches is, to a certain extent, blurred.}.

In the most common conditions, the state of the environment (and agent body) and the observations evolve over time in response to the agent's actions. In such cases, the state $z$ is often assumed to satisfy Markov conditions. In turn, due to typical sensory limitations such as limited field of view or occlusions, the environment is assumed to follow a partially observable Markov decision process (POMDP)~\cite{cassandra1994acting}. That is, when the current internal state of the environment is $z_t$ (where the subscript represents a discrete time step), performing an action $a_t$ causes the internal state to transition to $z_{t+1}$ and the corresponding $x_{t+1}$ is observed. In the POMDP case, the prediction and inference models are given as follows.
\begin{align}
\makebox[12em][l]{Prediction (transition):} &z_t\sim p(z_{t}|z_{t-1}, a_{t-1}) \nonumber\\
\makebox[12em][l]{Prediction (generation):} &x_t\sim p(x_t|z_t) \nonumber\\
\makebox[12em][l]{Inference:} &z_t\sim q(z_t|x_{1:t}, a_{1:t-1}),
\end{align}
where $x_{1: t}$ denotes the set of observations from the first step ($x_1$) to step $t$ ($x_t$). To learn a state space model (SSM) on the time interval $1$ to $T$,  a variational approach can be adopted that maximizes the following objective~\cite{hafner2019learning,hafner2019dream,hafner2020mastering}:
\begin{align}
&\log p(x_{1: T}|a_{1: T-1}) \nonumber\\
&\geq \sum_{t=1}^T \underbrace{\mathbb{E}_{q(z_t|x_{1: t}, a_{1: t-1})}[\log p(x_t|z_t)]}_{(Negative)\ prediction\ error} - \underbrace{\mathbb{E}_{q(z_{t-1}|x_{1: t-1}, a_{1: t-2})}[D_{\rm KL}[q(z_t|x_{1: t}, a_{1: t-1})||p(z_{t}|z_{t-1}, a_{t-1})]]}_{Regularization} \equiv \sum_{t=1}^TL_t,
\label{elbo}
\end{align}
where the first term in Eq.~\eqref{elbo} represents the prediction error (or reconstruction error) of the observation, and the second term represents the regularization for the state representation (so that the transition model and the inference model yield the same state representation).

\subsection{Predictive coding and the free-energy principle}
The original predictive coding model provided by Rao and Ballard~\cite{Rao1999} was proposed as a model of visual processing in the brain. The model assumes a hierarchically organized neural network, and top-down and bottom-up interactions at each hierarchical level are considered. In the top-down process, higher levels generate predictions about lower-level neural activities, and the lowest level generates sensory predictions. In the bottom-up process, residual errors between the predictions and actual activities (or sensory inputs) are computed and used to correct the originally generated predictions at each level. Predictive coding models learn spatial and temporal statistical regularities at each level for efficient coding and to reduce the redundancy of the predicted activity of lower levels \cite{huang2011predictive,hogendoorn2019predictive}. The main principle behind this hierarchical predictive coding cortical organization in the brain is prediction error minimization (PEM) \cite{Friston2005,friston2009predictivecoding,Friston2010}. This idea based on the principle of PEM has been extended to various cognitive processes, and this framework is usually referred to as predictive processing \cite{clark2013a,clark2015surfing,hohwy2020new}. This approach is being recently used also in machine learning to learn robust generative models of data \cite{ororbia2022neural}.

The principle of PEM can be situated within a more general principle of free-energy minimization because the amount of variational free energy, the core information measure used in the FEP, can be understood, under simplifying assumptions, as the amount of prediction error \cite{Friston2005, Friston2006b, Friston2010}.
The variational free energy $F_{t}$, which is an upper bound on the surprise $-\log p(x_t|x_{1:t-1}, a_{1:t-1})$, is the negative value of the evidence lower bound (ELBO) $L_{t}$ introduced in Eq.~\eqref{elbo} as follows.

\begin{align}
-L_{t} &= F_{t}\nonumber \\
&=\underbrace{-\mathbb{E}_{q(z_t|x_{1: t}, a_{1: t-1})}[\log p(x_t|z_t)]}_{Prediction~error}+\underbrace{\mathbb{E}_{q(z_{t-1}|x_{1: t-1}, a_{1: t-2})}[D_{\rm KL}[q(z_t|x_{1: t}, a_{1: t-1})||p(z_{t}|z_{t-1}, a_{t-1})]}_{Regularization}\nonumber \\
&=\underbrace{D_{\rm KL}[q(z_t|x_{1: t}, a_{1: t-1})||p(z_{t}|x_{1: t}, a_{1:t-1})]}_{Divergence} - \underbrace{\log p(x_t|x_{1:t-1},a_{1:t-1})}_{Evidence}\nonumber \\
&\geq \underbrace{-\log p(x_t|x_{1:t-1},a_{1: t-1})}_{Surprise}.
\label{vfe}
\end{align}
From the second line of Eq.~\eqref{vfe}, when observations are assumed to follow a Gaussian distribution with a fixed variance, minimizing the variational free energy is equivalent to minimizing the sum of mean squared errors and a regularization term. 

The FEP is a mathematical formulation of how self-organizing systems, such as biological agents, brains, and cells, are able to maintain an equilibrium with their environment by means of minimizing variational free energy, or the surprise associated with sensations\footnote{In contrast to states of surprise, free-energy can be measured because it is a function of sensory input and the inferred state \cite{Friston2010}} \cite{friston2011action,friston2009predictivecoding,buckley2017a}.
In the FEP, different cognitive processes such as perception and action can be understood as different ways to minimize the variational free energy in terms of probabilistic inference called active inference \cite{Friston2010e, Friston2012f, friston2015active} as detailed in the next subsection.

\subsection{Active inference and exploration}
Active inference is a normative framework that derives from the FEP and provides a unifying account for perception, control, and learning in terms of minimization of the variational free energy in the past, present, and future. This unification is important in neuroscience as it reflects neural mechanisms and on a computational level because it offers new perspectives and the possibility of sharing algorithmic solutions between all these functions and transforming them into sophisticated robotic behaviors~\cite{lanillos2021active}.
For example, perception aims to minimize the variational free energy in the past and present by inferring the latent representations of observed sensory inputs\footnote{This perception can be regarded as a variational Bayesian version of the original predictive coding that employs a maximum a posteriori (MAP) estimation \citep{friston2009predictivecoding}.}, e.g., when an orange appears in the field of view instead of the apple as currently encoded in the internal representation, the representation state can change toward that of an orange ~\cite{Friston2005}. Conversely, actions try to minimize the variational free energy in the present by actively sampling sensory inputs, e.g., by moving the gaze away from the orange toward an apple. In addition to selecting an action in the present, agents can infer a sequence of future actions (or policy) that elicit the most plausible future states ~\cite{Friston2012f,friston2015active,kruglanski2020a,friston2012optimal} by considering the minimization of \textit{expected} free energy, as detailed below.

While active inference introduces an important perspective towards an understanding of adaptive and autonomous behaviors, an obvious  behavioral imperative, the exploration-exploitation dilemma, seems in conflict with this idea because exploration, i.e., observing an uncertain aspect of the environment, would result in obtaining an unpredictable outcome \cite{friston2012free,sun2020dark}. 
Indeed exploration and active perception have a central role in robot control and learning. Several tasks focus on robots’ ability to explore an unknown environment \cite{chaplot2020learning,ramakrishnan2019emergence,Ammirato2017}. Furthermore, in social contexts, unobservable factors such as others' intentions must be actively considered to allow for efficient human-robot collaboration \cite{ognibene2013towards,lee2015stare,donnarumma2017action,Wang_2020_CVPR}. 

However, in \cite{friston2015active,schwartenbeck2019computational,Parr2022}, the authors showed that active inference can easily support exploratory behaviors and that it can provide an elegant formal solution for the exploration-exploitation dilemma.

In fact, we must consider that planning behaviors for an extended period of time require anticipating future data. More specifically, to infer the best action sequences (policies), one must also predict the  future observations they would produce. This is realized in the active inference framework by minimizing the expected free energy over a  time interval $T$. 
We can express this as the sum of two terms, including \textbf{i)} the variational information gain term \cite{Denzler2002Infotheory,Sommerlade2008InfoTheory,ognibene2013towards,hafner2022DivergenceMinimization}, or \textit{epistemic} value \cite{friston2015active}, defined as the expected KL divergence between the distribution of the latent states conditioned on the expected observations $q(z_{t+1:T}|x_{t+1:T}, a_{t:T-1})$ and the prior distribution on the latent states $q(z_{t+1:T}|a_{t:T-1})$ that represents the reduction in uncertainty on the latent states $z_{t+1:T}$ provided by the expected observations $x_{t+1:T}$, and \textbf{ii)} the extrinsic or \textit{pragmatic} value $\log p(x_{t+1:T}|C)$, where $C$ denotes the agent's preferences. 
This results in the following expression\footnote{Note that in the literature on active inference, a sequence of actions $a_{t:T-1}$ is referred to as a policy $\pi$. This policy is different from a policy in reinforcement learning, where it represents a statistical mapping from states to actions ($\pi(a_t|s_t)$). Using this notation of the policy $\pi$ and mean-field approximation, the general formulation of the expected free energy can be described by the following more practical formulation.  $G(\pi) = -\sum_{\tau=t+1}^{T}\mathbb{E}_{q(x_{\tau}|\pi)}[D_{\rm KL}[q(z_{\tau}|x_{\tau}, \pi)||q(z_{\tau}|\pi)]]-\mathbb{E}_{q(x_{\tau}|\pi)}[\log p(x_{\tau}|C)]$, where the time step $\tau>t$ used here is a future time step.}.
\begin{align}
G(a_{t:T-1}) =-\underbrace{\mathbb{E}_{q(z_{t+1:T}, x_{t+1:T}|a_{t:T-1})}[D_{\rm KL}[q(z_{t+1:T}|x_{t+1:T}, a_{t:T-1})||q(z_{t+1:T}|a_{t:T-1})]]}_{Epistemic\ value} \nonumber \\
    -\underbrace{\mathbb{E}_{q(x_{t+1:T}|a_{t:T-1})}[\log p(x_{t+1:T}|C)]}_{Pragmatic\ value}. \label{eq:expected_F}
\end{align}
The epistemic value term favors obtaining observations that disambiguate the world state such as obtaining the address for the best apple shop in town, versus observations that correspond to multiple (aliased) state such as corridors in a mall. Without the factor of variational information gain, asking the address of the shop would not be preferred to any other action that would not immediately result in obtaining an apple.  
Thus, minimizing expected free energy corresponds to maximizing the sum of epistemic and pragmatic values over an extended period and defines the optimal trade-off between exploration and exploitation. The similarity between the epistemic value term in Eq.~\eqref{eq:expected_F} and the divergence term in Eq.~\eqref{vfe} with an inverted sign may be noted. This is due to the different role that observations play in expected free-energy formulation, where they comprise not observed data but expected observations. Finally, the close connection between variational free energy (Eq.~\eqref{vfe}), expected free energy (Eq.~\eqref{eq:expected_F}), used in this context to define behaviors with exploration capabilities, and  ELBO (Eq.~\eqref{elbo}), used  to model learning processes objectives, shows the versatility of this type of formulation, the extension and refinement of which currently a promising field of research that aims to develop an autonomous system with the ability to efficiently acquire and execute complex skills \cite{hafner2022DivergenceMinimization}. For a more advanced and detailed presentation, we refer to \cite{friston2015active,parr2019generalised,hafner2022DivergenceMinimization,Parr2022}.

Another important framework that considers behaviors as inference is planning or {\it control as inference} (CaI)~\citep{attias2003planning,toussaint2009robot,kappen2012optimal,botvinick2012planning,millidge2020relationship}. The main difference between CaI and active inference is that CaI introduces a binary optimality variable $O_t$ that represents whether an action $a_t$ in state $z_t$ is optimal (or preferred)~\cite{van2020controlled,millidge2020relationship}. If the reward for taking action $a_t$ in state $z_t$ is $r\left(z_t, a_t\right)$, the conditional distribution of the optimality variable is defined as follows.
\begin{align}
    p\left(O_t=1 \mid z_t, a_t\right)\equiv\exp \left(r\left(z_t, a_t\right)\right).
\end{align}
Thus, unlike active inference, CaI can introduce the value of the reward at each time explicitly and independently of the observation's generative model\footnote{Therefore, unlike active inference, CaI does not require the assumption of POMDP.}.

CaI aims to obtain the optimal policy $p(a_t|z_t)$ for inference. If the variational inference is chosen as a solution to the intractability of exact inference (as with active inference), we seek the policy that maximizes the following ELBO\footnote{Here, MDP is assumed, i.e., state $z_t$ is an observed variable rather than a latent variable; if POMDP is assumed, inference and generative models for observations are added to this ELBO.}.
\begin{align}
    \log p(O_{1:T}) \geq \mathbb{E}_{\prod_{t=1}^T p(a_t|z_t)p(z_{t}|z_{t-1},a_{t-1})}\left[\sum_{t=1}^T r\left(z_t, a_t\right)+H\left(p\left(a_t \mid z_t\right)\right)\right],
\end{align}
where $H$ represents the entropy.
This corresponds to the entropy-regularized expected reward, and reinforcement learning with this as the objective is called entropy-regularized reinforcement learning~\citep{haarnoja2018soft}.

\section{Prior works}
\subsection{World models and model-based reinforcement learning in AI and robotics}
In this section, we describe world models used in \emph{model-based reinforcement learning} in the context of artificial intelligence and robotics.

\begin{figure}[tb]
\centering
\includegraphics[width=1.0\linewidth]{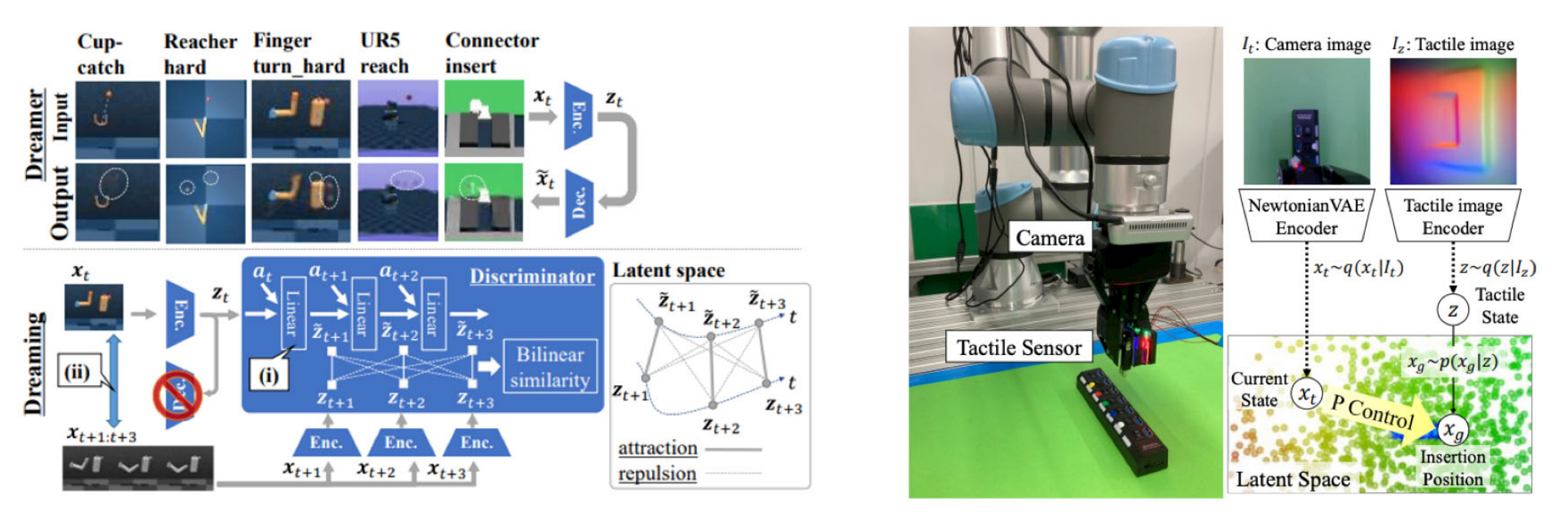}
\caption{Researches of world models. Left: Dreamer~\cite{hafner2019dream} and Dreaming~\cite{okada2021dreaming}. Right: robot control system using NewtonianVAE~\cite{Okumura2022}.
}
\label{fig:worldmodels_researches}
\end{figure}

Time-series world models conditioned on behavior have been studied for policy learning for some time.
Schmidhuber proposed learning an agent's policy (utility) via an RNN-based world model obtained by self-supervised learning~\cite{schmidhuber1990making}. Based on this idea, Ha et al. introduced a large-scale world model consisting of a VAE and an RNN that learned directly from observations (time-series images) from the external world~\cite{ha2018world}. They showed that the policies of agents trained only on this world model, which learns a game environment, can work properly in real game environments.
Since this study, research has been conducted on self-supervised learning of models of an environment directly from observations, along with ideas referred to as ``world models''.  However, the authors trained spatial compression (VAE) and temporal transitions (RNN) separately; thus, the perspective of learning state representations was not considered.

Subsequently, VAE-based models that simultaneously learn time transitions and spatial compression have been proposed.
Kaiser et al. proposed a VAE-based world model with discrete latent variables designed to predict the next frame and reward from the stacked frames of the previous four steps and the current action and showed that model-based reinforcement learning using this model performed adequately in an Atari video game environment  with high sample efficiency~\cite{kaiser2019model}.
One limitation of this model is that it does not include RNNs and cannot account for long-term prediction. Moreover, the measures are learned from the observation space, so the learned representation is not fully exploited. 
Ke et al. showed that learning long-term transitions using a stochastic RNN-based world model contributes to high performance on tasks that require long-term prediction~\cite{ke2019learning}.
All of these models, however, are autoregressive, requiring the generation of a high-dimensional observation space every step for long-term prediction, and are unable to transition within the latent space.

Recently, models that learn transitions in latent space without requiring autoregressive generation have been widely used.
Hafner et al. introduced a recurrent state space model (RSSM) that includes RNNs in SSM and showed that it could be used for long-term prediction and model-based reinforcement learning with higher performance than model-free learning~\cite{hafner2019learning}.

While PlaNet~\cite{hafner2019learning}, the first study using RSSM, used an existing model-based planning method (cross-entropy method) for planning in the latent space\footnote{PlaNet was extended to be uncertainty-aware on the basis of Bayesian inference~\cite{okada2020planet}.}, Dreamer~\cite{hafner2019dream}, a subsequent method, explicitly modeled the policy and value function in neural networks and learned a world model through gradients in an actor-critic framework, resulting in a better performance than PlaNet.
This model has been further developed by replacing the latent variables with discrete values, which significantly outperformed model-free performance in Atari game environments (Dreamer V2~\cite{hafner2020mastering}), and by using contrastive learning instead of reconstruction, which resulted in higher performance on tasks that were difficult to reconstruct (Dreaming~\cite{okada2021dreaming}, see the left side of Fig.~\ref{fig:worldmodels_researches}). They were also combined and compared (Dreaming V2~\cite{okada2022dreamingv2}).

In terms of obtaining a good state representation for control, enforcing explicit constraints on transitions is preferable. For example, NewtonianVAE was able to form PD-controllable state space~\cite{jaques2021newtonianvae}. However, to develop such a model, what kind of state representation the world model should acquire (as a good representation for control) should be considered, which remains as yet relatively unclear (see section 4.1 for details).

These world models have been shown to be effective in learning using real robots. Okumura et al. successfully applied the NewtonianVAE to a robot and enabled it to perform a precise socket insertion task~\cite{Okumura2022}(see the right side of Fig.~\ref{fig:worldmodels_researches}). Wu et al. showed that Dreamer V2 enabled real robots to perform online learning with very high sample efficiency and performance, which includes a pipeline of acquiring data through interaction with the external world, learning a world model, and controlling the robot using the model~\cite{wu2022daydreamer}.  However, all of these results are for a single environment and task, and what kinds of world models should be acquired for robot control in diverse environments and tasks remains unclear.

\subsection{Predictive coding and active inference in cognitive and developmental robotics}
In recent years, an increasing body of research has considered predictive coding models for perception and action in robotics. Recent comprehensive reviews on active inference and predictive processing in robotics can be found in~\cite{lanillos2021active,ciria2021predictive}, respectively. These ideas aim to provide a general mathematical account of behavior. Importantly, they incorporate adaptation and robustness to current methods in cognitive and developmental robotics.

\begin{figure}[tb]
\centering
\includegraphics[width=1.0\linewidth]{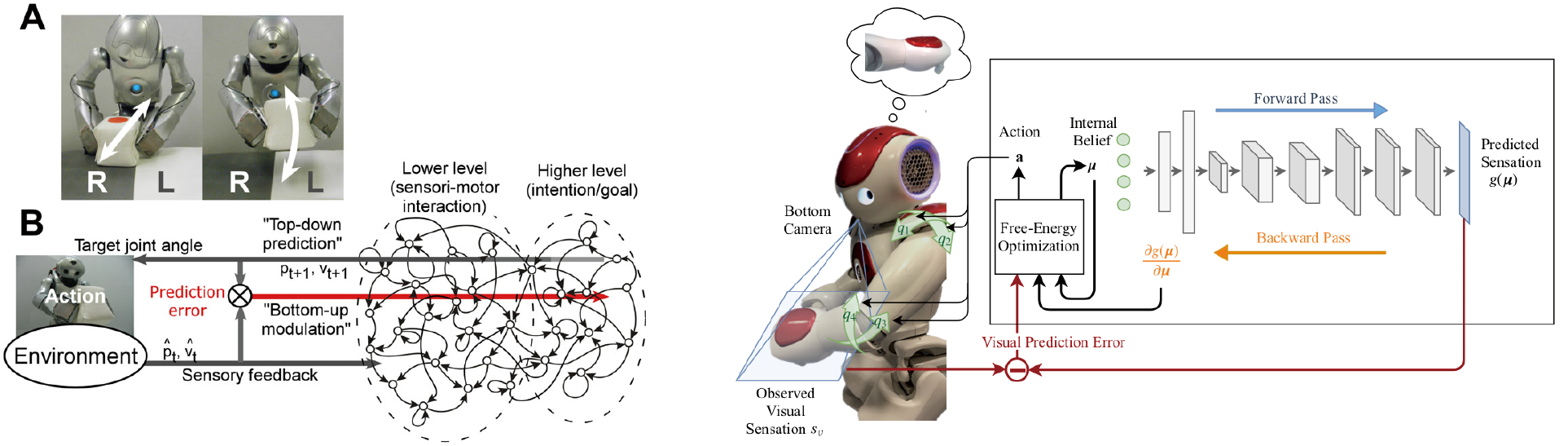}
\caption{Research on predictive coding and active inference in cognitive robotics. Left: adaptation to environmental changes by prediction error minimization~\cite{Yamashita2012a}. Right: body perception and action by pixel-based deep active inference~\cite{sancaktar2020end}.
}
\label{fig:PC_AI_robots}
\end{figure}

Since the early works of Tani et al. using hierarchically organized RNNs~\cite{Tani2003}, a variety of methods have been proposed to exploit this idea of prediction-error-minimization or propagation. ``Higher levels'' (internal representation) generate  predictions about the dynamics of the ``lower levels'' up to the sensorimotor level. Prediction errors at the sensorimotor level, given the observations, are then propagated ``upwards'' in the hierarchy correcting the internal state and thus minimizing the errors. 
Extensions of Tani's approach allow multiple time scales~\cite{Yamashita2008, Nishimoto2009, Namikawa2011, Yamashita2012a} (see the left side of Fig. \ref{fig:PC_AI_robots}), stochasticity~\cite{Murata2013, Murata2015a} and stochastic latent representations~\cite{Ahmadi2019}. In particular, a precision-weighting mechanism for the PEM enabled robots to extract stochastic or fluctuating structures of temporal sensorimotor sequences and utilize the extracted structures for their action generation~\cite{Murata2015a}. Interestingly, this mechanism is related to the precision account in psychiatric disorders~\cite{lanillos2020review} (especially autism spectrum disorder~\cite{VandeCruys2014, Lawson2014}) and several works have proposed cognitive robot models based on aberrant-precision to model unusual perception and action~\cite{Idei2018, Idei2020, Idei2021a}.

Aside from hierarchical RNNs, active inference controllers for robotic manipulators ~\cite{pezzato2020active,meo2021multimodal} and humanoid robots~\cite{oliver2021empirical,sancaktar2020end} have also been developed based on Lanillos's initial work on predictive coding adaptive perception and learning~\cite{lanillos2018active,lanillos2018adaptive} for both low-dimensional and high-dimensional inputs (see the right side of Fig. \ref{fig:PC_AI_robots}). These methods have widespread applications such as object manipulation~\cite{Ito2006a}, imitation~\cite{Ito2006a}, language acquisition~\cite{Sugita2005}, social interaction~\cite{Chen2016a} and navigation~\cite{ccatal2021robot}.

Cognitive robots benefit from predictive coding mechanisms to infer others' actions \cite{imre2019}. The reuse of common circuits for both movement generation and action estimation seems to be a key principle in the sensorimotor organization. Recently, the authors of \cite{seker2022} proposed deep modality blending networks (DMBN) designed to create a common latent space from the multi-modal experience of a robot by blending multi-modal signals with a stochastic weighting mechanism. Using a state-of-the-art skill-encoding system referred to as Conditional Neural Movement Primitives (CNMPs) \cite{seker2019rss}, they showed that deep learning could facilitate action recognition and produce structures to sustain anatomical (mirror-like) and effect-based imitation capabilities when combined with a novel modality-blending scheme.

Current state-of-the-art research is focusing on scaling active inference in planning tasks~\cite{buckley2017a} with high-dimensional inputs~\cite{van_der_Himst_2020,tschantz2020scaling} and improving representation learning through multimodal common latent space~\cite{seker2022} or introducing structural inductive biases, such as objects~\cite{van2022object}. Whilst active inference is a promising framework for robotics~\cite{da2022active}, current works are still limited to a particular aspect of cognitive and developmental processes. Therefore, in addition to extending the scalability of computational frameworks, continual or lifelong learning for developing abilities from low-level sensorimotor skills to higher-order cognitive functions should also be considered.

\section{Frontiers and Challenges}
\subsection{Latent representations for action planning}\label{sec:latent-rep}
\label{sec:representationsForAction}
One of the most important challenges in world-model approaches of any kind is that of efficiently performing planning, in the sense of generating meaningful actions to solve a sequential task~\cite{valenzo2022grounding}. Working in the high-dimensional space of the sensorimotor manifold is very computationally expensive and provides local optima solutions~\cite{sancaktar2020end}. In fact, current approaches in planning use a compressed encoded representation of the world dynamics, which aids in the process of predicting future states and in action generation~\cite{ha2018recurrent}. In reinforcement learning, state representation learning is tied to learned tasks to achieve high performance because it depends on the actions needed to obtain the maximum expected reward~\cite{hafner2019dream}. However, this sometimes prevents generalization across tasks. Decoupled action-representation world models are an interesting work-around~\cite{laskin2020curl}. In deep active inference~\cite{lanillos2021active} amortized methods have also been considered~\cite{van_der_Himst_2020,fountas2020deep}, in addition to contrastive~\cite{mazzaglia2021contrastive} and iterative amortized inference approaches~\cite{van2022object}. 

However, the key question cannot be narrowed down to that what type of architecture or method should be used. Rather, what type of information should be encoded in the latent representation and how this information is processed must be a key focus so that information is not uncoupled from the sensorimotor process, particularly from motor control, which is a key limitation of existing endeavors in robotics.
There has been considerable discussion as to what would comprise an appropriate state representation of a world model; that is, what inductive bias or prior knowledge should be given~\cite{bohmer2015autonomous, achille2018separation, lesort2018state}. Here, we list the properties of this prior knowledge we consider important. 
\begin{itemize}
      \item {\bf Low dimensionality.} Observations obtained from the environment are high-dimensional, and compressing this information into a low-dimensional space is critical for efficient data handling, abstraction, and planning. This approach is the most frequently considered in state representation learning. The challenge is how best to represent observations in a low-dimensional encoding while retaining the necessary task-dependant information. Recent literature focused on generative and discriminative approaches to tackle this.
      \item {\bf Meaningful abstraction and disentanglement.} Low-dimensional representations should have scene-understanding and task meaning, such as objects~\cite{van2022object}, locations~\cite{stoianov2022hippocampal} and temporal events~\cite{gumbsch2021sparsely}. Representation disentanglement proposes that factors of variation with different semantics should be separated, contributing to the requirements for sufficiency and efficiency in state representation. Object-centric representation learning is related to this hypothesis, \cite{greff2019multi,van2022object}, in which every observed object is encoded independently. 
      \item {\bf Compositionality.} Although disentanglement aims to separate independent factors, the agent should also acquire their relationships and hierarchy. In the case of object representations, there should also be relations or implication relations among objects. Currently, methods such as those using graph neural networks are being considered, but they do not provide an essential solution. This idea of the compositionality of representation is also relevant to the neuro-symbolic approach.
      \item {\bf Dynamics prediction.} These three properties are important not only for learning representations in static environments but in dynamic worlds, e.g., they consider a transition model that depends on external factors and agent actions. The best latent representation is one that allows transitions to be easily predictable for given actions. Many recent models use RNNs to learn transitions, which incorporate information on long-term dependence~\cite{hafner2019learning,hafner2019dream}. One way to make transitions more predictable is to incorporate prior knowledge of the physical world (e.g., dynamics following Newton's laws of motion~\cite{jaques2021newtonianvae}). Furthermore, by learning to separate representations that are not related to control from state representations, representations that are easier to control can be acquired~\cite{wang2022denoised}.
      \item {\bf Values are sufficiently encoded.} To perform reinforcement learning on the state representation of the world model, the value of the state representative to the agent must be known. For example, a recurrent state representation learns to predict the reward from the state so that the reward is embedded in the  representation~\cite{hafner2019dream}. However, because the value of the state changes depending on the task, it remains unclear whether this hypothesis should be introduced in a world model that should acquire a prediction model that is as task-independent as possible. Alternatively, in active inference approaches, the agent value function cannot be modified, and it is defined by expected free energy. Here, the challenge becomes learning the state preferences and being able to predict the transitions that may yield those preferences.
      \item {\bf Task-agnostic.} Representations should be informative to solve narrow problems where the agent is trained but also sufficiently general to be reused in tasks with different kinds of variability or new tasks that the agent has never encountered.
      \item{\bf Fusion of multiple-types multimodal information.} Robots inevitably face a variety of events with their multimodal sensorimotor systems. Observations given to world models are from multiple sources (e.g., social non-social, sensorimotor purely sensorial, and linguistic and non-linguistic). They can have different reliability and volatility and represent various aspects of the world. Therefore, the world models must properly encode the internal representation in a stable and efficient manner.
      \end{itemize}
These are some of the elements that we identified that a latent representation should be fulfilled to provide a smooth connection with real-world interaction and provide power for solving cognitive tasks. Importantly, abstract representation and disentanglement, such as objects or events encoding, may be important to achieve efficient planning, reducing the gap for neuro-symbolic solutions. However, the connection between the low-dimensional (and hierarchical) encoding and the synchronization with the sensorimotor control remains a major challenge. %In the following subsection, we further explore neuro-symbolic world models.

\subsection{Neuro-symbolic predictive models}\label{sec:neuro-symbolic}
In this section, we provide an overview of state-of-the-art techniques in which symbols and rules are discovered and used by robots through neuro-symbolic approaches. The term {\it symbols}, here, refers to manipulative discrete representations used in symbolic AI and cognitive science. The neuro-symbolic approach attempts to integrate conventional symbolic and modern neural network-based AIs.

Both biological and artificial agents benefit from predictive coding mechanisms for reasoning, decision-making, and planning. Predictive forward models are used to generate plans that involve a sequence of actions. For example, chimpanzees are known to generate multi-step plans that include stacking a number of boxes on top of each other, grabbing a long stick, climbing on top of a stack of boxes, and using the stick to reach the object that was initially out of reach \cite{Kohler1927,Steedman2002}. While the underlying cognitive mechanisms for high-level planning remain unknown, different specific brain regions have been shown to become active in inductive and deductive reasoning in humans \cite{Goel1998} while predicting the effects of actions \cite{Al2014}. In artificial agents, on the other hand, standard search and planning rely heavily on manually coded or learned state transitions and prediction models \cite[Ch.\ 3–6,10-11]{Russell2016}. 

The seminal work of \cite{mugan2012autonomous} addressed the learning of discrete representations of predictive models, i.e., dynamic Bayesian networks, by discretizing the continuous features of the environment to plan goal-directed arm/hand control. \cite{ahmetoglu2022high} showed the units generated by slow feature analysis with the lowest eigenvalues resemble symbolic representations that highly correlate with high-level features, which might be considered precursors for fully symbolic systems. \cite{Konidaris2014,Konidaris2015} studied methods to discover useful symbols that can be directly utilized in problem and domain definition language (PDDL) for various agent settings. In simulation and the real world, the discovered symbols were directly used as predicates in the action descriptions to generate deterministic and probabilistic symbolic plans. \cite{james2019learning} learned symbols in the ego-centric frame of the agent to transfer the learned symbols into novel settings. \cite{Ugur-2015-ICRA,Ugur-2015-Humanoids} discovered symbols in the continuous perceptual space of the robots for PDDL-based manipulation planning via combining several machine learning algorithms such as X-means clustering and support vector machine (SVM) classification. Although the symbols were discovered by the robot without any human intervention, the continuous perceptual features were manually encoded by the authors. Towards an end-to-end framework, \cite{Ahmetoglu2022} used directly raw camera image and pixel values to discover symbols via a novel deep predictive coding neural architecture. In detail, they proposed a deep encoder-decoder network with a binary bottleneck layer designed to take a camera image and an action as input and output the action effects in pixel coordinates. The binary activations in the bottleneck layer encode object symbols that not only depend on the visual input but are also shaped based on action and effect. In other words, the objects that provide the same affordances \cite{Sahin2007} were automatically grouped together as object symbols. To distill the knowledge represented by the neural network into rules useful for symbolic reasoning, a decision tree was trained to reproduce its decoder function. Probabilistic rules were extracted from the effect predictor / neural decoder and encoded in the probabilistic PDDL, which can be directly used by the off-the-shelf AI planners. In follow-up work, \cite{Ahmetoglu2023} used a multi-head attention mechanism to learn symbols to encode affordances of a varying number of objects.

Asai et al. implemented a neural framework where a state autoencoder with a discrete bottleneck layer was trained first, and preconditions and effects of actions were learned next~\cite{asai2017classical}. In follow-up work, \cite{asai2020learning} combined the previous two systems and discovered  action preconditions and effects together with visual symbols. These works were realized in visual environments such as 2-D puzzles and achieved visualized plan executions. An important aspect of pure visual neuro-symbolic systems and studies on neuro-symbolic robotics is that in robotics, predictive coding over object symbols takes actions and effects into account in addition to the features of objects and the environment, which facilitates the formation of symbols that are likely to capture object affordances \cite{Gibson1979,Zech2017,jamone16aff,renaudo2022computational}.

However, in the context of world models, methods to integrate prior symbolic knowledge into VAE and SSM-based world models are still being explored. The bottom-up formation of symbolic representations is closely related to disentanglement and compositionality discussed in Section 4.1. Moreover, leveraging linguistic knowledge in world modeling is a challenge in relation to neuro-symbolic predictive models.

\subsection{Affordance perception}\label{sec:affordance}
Affordance perception has often been discussed independently of world models, but in fact, it is closely related. According to the original definition provided by Gibson~\cite{gibson66aff,gibson77aff,Gibson1979}, an affordance is an action possibility offered to the agent by the environment. A stable surface may afford to be traversed; a stone may afford the possibility of being used as a hammer; a door handle may afford the possibility of being opened. The concept of affordances and affordance perception has then been further analyzed and revisited in psychology, neuroscience, cognitive science, artificial intelligence, and robotics (see \cite{jamone16aff} for a recent survey). The ability to perceive affordances is crucial for any biological or artificial agent to interact successfully with the environment.

Central to the idea of affordances is that the action possibilities depend on both the agent and the environment; the same environment would offer different action possibilities to different agents, depending on their sensorimotor capabilities. A stable surface affords the possibility of traversal to an agent that is able to locomote and whose body dimensions fit the size of the surface borders; a stone affords the possibility of being used as a hammer to an agent who is able to pick it and who has enough force to lift it; a door handle affords to open to an agent who knows how to open doors, assuming it is well-designed \cite{norman99aff}. Therefore, those affordances must be \emph{learned} autonomously by the agent. In fact, the agent must learn how to perceive them. The means by which agents perceive affordances are those of ecological perception, powerfully illustrated by Eleanor Jack Gibson \cite{gibson94aff,gibson00aff,gibson03aff}: ``narrowing down from a vast manifold of (perceptual) information to the minimal, optimal information that specifies the affordance of an event, object, or layout’’ \cite[p.284]{gibson03aff}. The agent must learn what minimal information is to be picked; this happens both through evolution and development, leveraging the sensorimotor exploration of the environment by physical interaction.

Interestingly, while exploring the action possibilities, the agent can learn the effects of those actions as well. This is crucial for biological agents and turns out to be extremely useful for artificial systems as well. In fact, most computational models of affordances in robotics rely on representations that include not only the action but also the effects (or in other terms, the goal) of the action \cite{sahin2007afford,montesano08aff,Ugur2011,krueger11aff,Szedmak2014,goncalves14aff,dehban16aff,dehban17aff,ugur17aff,sarathy18aff,strama18aff,imre2019,seker2019rss,tekden2020}; therefore, the perceived possibilities for actions (and for achieving certain effects) can be used for action planning, leading to problem-solving \cite{antunes16planning}. 
Such comprehensive models of affordances are, in fact, world models; they are internal models of how the world behaves ``in the eyes'' of the learning agent, and they can be used by the agent to make predictions about how the world will change if certain actions are performed. Therefore, it is not surprising that the computational techniques used for learning affordance models often overlap with those used for learning world models \cite{dehban22aff}.
It is worth noting that, in world model approaches, a robot only receives raw sensory information and needs to extract the relevant semantics from such data flow; therefore, to successfully integrate affordance perception in these systems, the challenges of meaningful abstraction/disentanglement and object-centric representation learning, described in Section 4.1, are particularly relevant.

\subsection{Social interaction}\label{sec:social}
Robots' ``worlds'' do not consist of physical objects alone but also of social entities, i.e., people who give them social guidance and try to cooperate with them. World models should model and predict social dynamics involving people's behaviors, and infer their latent variables, e.g., intentions and emotions, to cooperate with them, that is, to control social phenomena. 

Efficient and safe human-robot collaboration and interaction are some of  the main research objectives of robotics and have important practical applications \cite{shen2016investigation,albrecht2018autonomous,el2019cobot,hentout2019human,magrini2020human,ognibene2022active,semeraro2023human}. 
Associating beliefs, intentions, or mental states to other agents, theory of mind, or, in other words, trying to predict the internal state of another agent’s world model to understand its activities and context \cite{ognibene2013contextual}, is an essential aspect of human interaction \cite{fotopoulou2017mentalizing,veissiere2020thinking,saxe2006uniquely} and has attracted attention in robotics \cite{bianco2019tomadvantages}. 

Mutual understanding using a world model in social interaction can play an important role when complex interactions are challenging the perceptual systems of the agents, inducing a mismatch between their interpretation of the current context \cite{baker2017rational,bianco2022learningtom}. It is also crucial when different levels of knowledge and expertise induce different representations of a domain, as well as different points of view, which may induce conflictual interactions \cite{bianchi2021sweat} or different support strategies~\cite{ognibene2019proactive}. For example, the perspective of an automotive mechanic and that of an ordinary user differ considerably, so collaboration may be difficult if one cannot properly infer the internal state of others.
A robots’ world model can play a crucial role in its operation and functionality \cite{heinze2004modelling}.

The recent progress in machine learning methods has resulted in substantial improvement in action recognition methodologies \cite{roggen2010collecting,zeng2014convolutional,yang2015deep,Wang_2020_CVPR}. 
However, this approach has often focused on shallow and purely perceptual representations of the observed activities resulting in limited flexibility in terms of contexts, tasks, and observed actors demanding a substantial amount of difficult-to-collect data and retraining time to apply the system in relatively similar conditions \cite{lee2019model}. Approaches such as goal recognition as planning or inverse planning \cite{ramirez2009plan,baker2009action,sohrabi2016plan,albrecht2018autonomous}, that, given a model of the environment, understand others' activities by computing plans that would result in the observed actions have shown the flexibility advantage delivered in intention recognition by  a world model. 
Several works have extended this approach. The problem of dealing with behaviors generated under partial observability, which may require inferring both the plan and the beliefs, the mental state \cite{bianco2019tomadvantages,bianco2022learningtom}, of the observed actor, was studied with both classical planning \cite{ramirez2011goal} and Bayesian approaches \cite{baker2011bayesian,baker2017rational}. The impact of missing observations for the observer agent has also been analyzed~\cite{ramirez2011goal}. A further step has been proposed by active methods for activity recognition \cite{shvo2020active,amato2019active,ognibene2013towards} that use the same world model both to interpret others' actions as well as selecting actions that would improve the recognition process, e.g., by giving access to the most informative observations \cite{lee2015stare} and allow the completion of a joint task \cite{ognibene2019proactive}.
While even the initial formulations of this approach were computationally aware \cite{ramirez2009plan}, their efficiency is often affected by the length of the observed behavior and the environment complexity, resulting in methods that can seldom be applied online on a robot. Several models proposed a pre-compile approach that transformed the world into a form that would allow efficient plan recognition \cite{lee2019model}. The adoption of hierarchical world model representations has also been considered to constrain the computational and modeling costs of the process \cite{demiris2006perceiving,cardona2017toward,proietti2021active}. Precomputed and robust local plans, in the form of the same motor controllers that the robot uses to perform its own actions, have also been adopted to allow active perception for action recognition and prediction on humanoid robots \cite{ognibene2013towards,ognibene2013contextual}.
One of the main issues of the approach, also related to computational efficiency considerations, is relying on specific algorithms for planning that aiming for the optimal plan may misinterpret the bounded rational behaviors that collaborators may perform. 
This problem was faced by using online Bayesian inference in \cite{NEURIPS2020_df3aebc6}.

The additional flexibility provided by world models in social interaction skills is likely relevant beyond activity recognition. It is easy to imagine that purely supervised models may be limited in terms of perspective-taking and the ability to reason based on the world structure may help to adapt to partners with different sensory systems \cite{johnson2005perceptual,pandey2013towards,fischer2019computational,bianco2019tomadvantages}. Similarly, world models are likely to help with imitation learning by dealing with embodiment mismatch between the observed actor and the learner \cite{torabi2019recent}. Finally, physical cooperation \cite{mortl2012role,li2016framework} and signaling \cite{pezzulo2019body,ognibene2019implicit} would also be more flexible when integrating world and partner models in the equation, for example, to account for the trust of the human cooperator towards the robot \cite{kok2020trust}. Finally, a world model may also be learned through socially rich experiences and sources of information (e.g., imitation \cite{torabi2019recent} or verbal instructions) in addition to the results of autonomous exploration. However, developing a robust, efficient, and flexible enough representation may prove to be one of the main challenges in this effort.

\subsection{Brain-inspired world models}
In cognitive science, it has long been postulated that the brain learns small-scale models of the world and uses these models for various cognitive functions, such as perception, planning, and imagination~\cite{Craik1943}. For example, theories of perception-as-inference described perception as an inferential process, which works by ``inverting'' a generative model of how the percepts are generated \cite{Helmholtz1867,Gregory1997}. As discussed above, these ideas (and others) have been recently formalized under the label of the \emph{Bayesian Brain} \cite{Doya2007} and extended by \emph{Active Inference} from the domain of perception to other domains, such as action planning and interoception~\cite{Parr2022}. 

In parallel, there have been many attempts to describe mathematically and to assess the neuronal underpinnings of world models and of inference processes empirically (e.g., \cite{shiffrin2020brain}). One question that has received a great deal of attention is how the brain might encode internal world models in the neuronal substrate. Given that the brain models are often assumed to be probabilistic, various formal schemes have been proposed that describe plausible neuronal implementations of probabilistic variables and of Bayesian inference over these variables, such as, for example, probabilistic population codes \cite{Ma2006} and sampling schemes \cite{Buesing2011}. These attempts show that (probabilistic, generative) world models could be at least potentially implemented in neuronal substrate \cite{sebastian2020local,lynn2020humans} -- and even updated after statistical learning \cite{Berkes2011} -- but the specific scheme(s) that the brain might use for this remain to be fully assessed. 

Another relevant question is what algorithms the brain might use to perform inference over world models. A strong candidate in neuroscience is \emph{predictive coding} \cite{Rao1999,Friston2005}. Several studies have aimed to validate its key empirical predictions, showing that under the appropriate conditions, it is possible to observe predictions \cite{Ekman2017}, prediction errors \cite{Daw2011a} and other signatures of inference in brain signals \cite{Gomez2019} and that neural activity in lower visual areas in the absence of bottom-up inputs could be explained by the top-down, feedback dynamics postulated by predictive coding \cite{Muckli2015}. These and other studies (see \cite{Walsh2020} for a recent review) lend some support for predictive coding, but the theory remains under development. 

At yet another level, one may ask what the systems-level architecture that supports world models and whether different parts of the brain might model different aspects of the world is. Anatomical considerations suggest that the brain is not a monolithic entity but rather is composed of several areas and networks \cite{Bullmore2009}; however, the extent to which these areas or networks are modularized and how they exactly influence each other are heavily discussed \cite{Laird2017}. One interesting consideration is that cortical brain areas in humans and monkeys appear to be organized along principal gradients (defined by functional connectivity); in one of these gradients, heteromodal areas (e.g., prefrontal cortex) are placed at the top, and unimodal areas (e.g., primary visual area) at the bottom, recapitulating the structure of a putative hierarchical generative model \cite{Margulies2016}. Another interesting consideration is that there seems to be a ``division of labor'' between brain pathways that perform complementary computations, such as the two visual pathways for processing ``what'' and ``where'' information \cite{ungerleider94}. These anatomical and functional separations might be potentially interpreted as useful \emph{factorizations} of the brain generative models. 
A whole-brain probabilistic generative model (WB-PGM) approach attempts to build a cognitive architecture for cognitive and developmental robots integrating probabilistic generative model (PGM)-based modules referring comprehensive knowledge of human and animal brain architectures and their anatomy~\cite{taniguchi2022whole}.

The above studies indicate that at a general level, both neuroscience and machine learning / AI conceive world models and inference in similar ways. However, at a more detailed level, there might be profound differences between the ways these two disciplines use the same concepts. Predictive coding and other biological schemes proposed in neuroscience exploit top-down dynamics (and recurrences) in ways that are rarely used in machine learning. Furthermore, brain information processing is heavily based on spontaneous brain dynamics, which are largely absent in machine learning systems; see \cite{Singer2021,Pezzulo2021b,gyorgy2019brain} for a detailed discussion of putative computational roles of spontaneous dynamics. Moreover, it is plausible to assume that different parts of the brain might be specialized (or might have different \emph{inductive biases}) to process different statistical regularities, rendering them able to learn and model (for example) slower or faster dynamics of the visual scenes, one's own body, the actions of other agents, or extended temporal events \cite{Pezzulo2017a}. It is worth highlighting here that, although prediction errors have a central role in learning, there are other forms of statistical learning, such as those based on Hebbian associative learning \cite{nazli2022statistical}. It remains to be understood how to best endow our more advanced machine learning systems with the ability of the brain to perform (apparently) specialized computations but also orchestrate them coherently. Finally, it is important to remember that the brain is an evolved system, and our more advanced cognitive abilities are grounded in (the neuronal mechanisms supporting) simpler sensorimotor skills \cite{Pezzulo2016,Cisek2019}. Trying to develop advanced cognitive systems without the necessary requirements for embodied interaction and ``phylogenetic refinement'' might lead to solutions that differ completely from how the brain works -- or that fail altogether.

\subsection{Cognitive architectures}
Truly cognitive and developmental robots, i.e., embodied AGI,  that behave autonomously and flexibly in the real environment would have a wide range of sensors and exhibit multiple functions. That requires a large-scale world model that deals with multimodal sensory observations and multilayered state representations. Considering the discussion in Section 4.5, such word models may be factorized in a proper manner from engineering and biological viewpoints.  
To realize embodied AGIs, further frameworks and architectures to factorize a total world model into cognitive modules and to integrate individual cognitive capabilities into a cognitive system are required.
The idea is related to {\it cognitive architectures}, which have been studied in cognitive science, artificial intelligence, and robotics~\cite{lieto2018role,kotseruba202040}. 

In cognitive science, cognitive functionalities such as memory, perception, and decision-making are implemented as modules in the cognitive architectures studied, and the specific task can be solved by activating these modules coordinately. 
ACT-R \cite{anderson2009can} and Soar \cite{laird2008extending} are representatives of cognitive architectures. 
It has been shown that the model implemented by ACT-R can explain the time to solve the task by humans, and activation patterns of the brain can be predicted by activation patterns of the modules~\cite{anderson2013human}. 
Furthermore, Soar has been used for controlling robots~\cite{puigbo2013controlling} and learning games~\cite{mohan2009learning}. 
However, complex machine learning methods that have rapidly advanced in a decade are not introduced yet. 
Sigma \cite{rosenbloom2016sigma,damgaard2022toward} is a newer cognitive architecture that introduces the generative flow graph, a generalized probabilistic graphical model. 
Therefore, the model can be implemented using probabilistic programming techniques \cite{bingham2019pyro,paige2017learning,tran2016edward}. 
Furthermore, the concept of the standard model of the mind is discussed through a synthesis across these three cognitive architectures \cite{Laird2017}. Particularly, cognitive architectures based on first principles, e.g., with a general computation scheme, such as free energy minimization~\cite{Friston2005}, are especially attractive. 
The architecture for social cognition has also been proposed \cite{sandini2018social}. 
The authors point out that these architectures explained above are incomplete in dealing with the social aspect of cognition and describe the elements of architecture for social cognition. 
Clarion \cite{Clarion} is another cognitive architecture based on dual process theory \cite{kahneman2003perspective}. 
In this architecture, each subsystem is composed of explicit and implicit processes, and it is shown that the interaction between implicit-explicit processes can explain psychological phenomena. 

In robotics, several types of cognitive architecture have been proposed. One of them is ArmarX \cite{vahrenkamp2015robot}, which has three layers, including a middleware layer, a robot framework layer, and an application layer.  This three-layered structure simplifies the development robotics software easier.  
(Neuro-)Serket \cite{serket_nakamura18,taniguchi2020neuro} is another approach to integrating cognitive modules\footnote{Neuro-SERKET is an updated version of SERKET.}.
In (Neuro-)Serket, modules are described by the (deep) PGM and trained mutually by exchanging messages between modules. 
To make it easy to develop large-scale models, the modules in Neuro-Serket are weakly connected through the Serket interface.
(Neuro-)SERKET is closely related to the world model-based approach because SERKET requires each module to be a PGM, i.e., a model based on prediction and inference as Eq.~\eqref{eq_gm}, and integrate modules into a large PGM.   
This architecture does not provide any restrictions regarding the functionalities of modules. 
Therefore, it has high flexibility but brings high dimensional design space at the same time.
To reduce the large degree of freedom in the design space, a brain-inspired approach, WBA-PGM, was proposed~\cite{taniguchi2022whole}. 
In this approach, a cognitive model was constructed by connecting PGM-based modules utilizing knowledge from neuroscience. 
By referring to the brain studies, WBA-PGM constrains the function of modules and their connection and reduces the design space of the cognitive model.

There are two crucial requirements for cognitive architecture for cognitive and developmental robots, which can be used along with the approach based on world models and predictive coding. 
The first is the engineering aspect which is seen in (Neuro-)Serket and ArmarX. The scale of cognitive models that enables the robots to behave flexibly in the real environment is very large, and many modules must be connected and work collaboratively. 
Furthermore, the model needs to introduce machine learning techniques that are not only existing as well as those will be developed in rapid progress. 
The development of such a model would require a massive engineering effort, and this is considered a notable obstacle to realizing such robots.  
Therefore, architecture is needed to simplify development. 
Another requirement is that of the scientific aspect seen in ACT-R, Soar, WBA-PGM, and Clarion. Developing AGI, which is human-like intelligence, referring to the knowledge regarding humans obtained in cognitive science and neuroscience, can accelerate its development. 
However, meeting these two aspects completely is very challenging.
All machine learning techniques and module connections might necessarily be not reasonable from the point of view of cognitive science and neuroscience. 
On the other hand, entire humans are not understood yet. 
Therefore, finding common ground between engineering and science aspects and developing a novel cognitive architecture is a current challenge. 
Developing a large-scale cognitive architecture and overcoming the problems described in the previous subsections is also a challenge.

\section{Discussion}
As we described, world models and predictive coding are promising approaches in cognitive and developmental robotics. Before closing this paper, we will mention some remaining issues which have not been addressed in the main body sufficiently.

\textbf{Language and world models}:~Umwelts, i.e., worlds from first-person views, of biological systems are not monolithic but have some sort of structure. Notably, language and symbolic systems have syntactic structures. The interaction between high-level cognitive capabilities, e.g., language and reasoning, and low-level cognitive capabilities, e.g., perception and action, is essential in world modeling. Recently, large-scale language models (LLMs) have been replacing many natural language processing methods~\cite{devlin2019bert,brown2020language}, including reasoning tasks, which have been conducted solely by symbolic AI by the end of 2010s~\cite{wang2022self,kojima2022large}. Recently, the use of LLMs in robotics has been attempted, e.g., ~\cite{saycan2022arxiv}. 
It is clear that language learning and understanding by robots is itself a frontier~\cite{tangiuchi2019survey}. 
To leverage the symbolic knowledge in LLMs, integration of LLMs and world models will be an important challenge.

This shift from models of artificial symbols in conventional AI to models of natural language, i.e., a human symbol system, is resonating with the discussion in symbol emergence in cognitive and developmental systems~\cite{Steels2008,taniguchi2018symbol,Taniguchi2016SER}. An important topic is then considering not only the integration of human language into robots' world models in a top-down manner but also the bottom-up formation of symbol systems, including language in relation to world models.

\textbf{Policy representations}:~How should the policies of robots be represented? Conventionally, policies are described as feedback controllers $\pi(z_t, a_t) = p(a_t|z_t)$ in reinforcement and imitation learning. Even though one direction of world model approaches is to explore task agnostic representations (Section \ref{sec:representationsForAction}), the decomposition of world modeling and policy learning can be controversial. In a conventional approach of world models, policies ($\pi=p(a_t|z_t)$ or $a_{t:T}$) and world models ($p(z_{t+1}|z_t, a_t) $ and $p(o_t|z_t)$) are decoupled. In contrast, a series of studies about predictive coding in neuro-robotics have been intentionally entangling policies and world models and making robots directly learn $p(o_{t+1}, a_{t+1}| o_{1:t}, a_{1:t})$ and exhibiting many successful results in robotics, e.g.,\cite{jung2019goal,ito2022efficient,tani2016exploring}. 
As the notion of affordance also suggests, actions and perceptions are not independent and entangled, generally. The question ``to what extent should we decouple world model and policy representations?’’ should be investigated.

\textbf{From artificial cognition to human cognition}:~Cognitive and developmental robotics are also constructive approaches to human developmental cognition. Not only learning from neuro-, cognitive and developmental sciences but also provide with scientific feedback to them is also an important mission. Building a virtuous circle between studies on human and artificial studies is a challenge. 

The constructive approach may give us a novel approach to scientific and philosophical hard problems like self-awareness~\cite{hoffmann2021robot} and consciousness. The relationship between the multimodal world model and global workspace theory was suggested~\cite{juliani2022link}. Extending the discussion between world models and consciousness using robots may be an exciting challenge. Moreover, the relationship between predictive coding and emotion is worth exploring to build emotional robots and understand the emotions of biological systems~\cite{seth2013interoceptive,bauermeister2022role,hieida2022survey}.

\textbf{Software frameworks for implementation}:~To accelerate the studies on world models and predictive coding in robotics, the development framework for cognitive and developmental robotics is crucially important. In robotics, not only AI ``software’’ frameworks but also middle-ware are important. Recently, ROS has been widely used in the robotics community for bridging hardware and AI software layers. Developing and sharing such software frameworks as a community will be important, e.g., \cite{Lotfi2023sii}. Moreover, the world model involves many types of knowledge, and the knowledge can be used for achieving multiple functions via active inference. The software framework should allow the world model to efficiently organize the knowledge and to perform (cross-modal) active inference. A great initiative is the discrete state-space active inference python library~\cite{heins2022pymdp}. However, it can only be used for toy examples due to scalability issues and to be useful in cognitive and developmental robotics. It needs further development. For instance, the support for high-dimensional input observations and the possibility of combining discrete and continuous action and state representations are something that has been addressed in robotic approaches~\cite{lanillos2021active}.  

\textbf{Data-efficient and autonomous learning}:~A generalist agent called GATO was developed based on Transformers and shown to be able to solve various tasks with one neural network~\cite{reed2022generalist}. Although the approach is superficially different, the approach is really related to the world models and the predictive coding approach. However, the learning system is hugely data-hungry. It is very questionable if the model can be regarded as a model of human intelligence. Moreover, to train the generalist agent, researchers need to prepare a large dataset and simulation environment. 
Human children can autonomously explore their environment and acquire data through active exploration. Moreover, they use heuristics and biases in their developmental process. Learning and considering the human developmental process will give us the inspiration to build real generalist agents. 
Developing a data-efficient autonomous learning architecture with world models and predictive coding at its core is the key to a truly cognitive and developmental system.

\textbf{Emergence of behaviors}:~Should an agent have a completely internal model of its world? Lastly, we raise fundamental speculation about the world model-based approach. 
Behaviors are not externalization of internally designed trajectories but something to emerge through the interaction between the body and the environment.
For example, it has been proven by passive walking machines that the behavior of walking emerges only from the interaction between the body and the environment, without any computation by the brain~\cite{mcgeer1990passive}. About three decades ago, Brooks famously advocated the physical grounding hypothesis together with subsumption architecture, saying the world is its own best model~\cite{brooks1990elephants}. The robots behaved smoothly and flexibly without any explicit world models. This is also referred to as morphological computation, which means the body itself implicitly processes information dynamically~\cite{pfeifer2007self,pfeifer2009morphological}. Soft robotics emphasizes these points nowadays. Combining the viewpoints of the emergence of behaviors with complex physical dynamics and the world model-based approach is another important challenge.

\section{Conclusion}
In this survey paper, we have aimed to clarify the frontiers and challenges of world models and predictive coding in cognitive and developmental robotics. 
Creating an autonomous robot that can actively explore the real environment, acquire knowledge, and learn skills continuously is the ultimate goal of cognitive and developmental robotics.
To make the robot continuously develop through active exploration, the robot's learning process should be based on sensorimotor information obtained through physical and social interactions with the physical and social environment. Following the motivation, this paper reviewed studies related to world models and predictive coding in cognitive and developmental robotics and related AI studies. 
We clarified the definition of world model and predictive coding in robotics, in conjunction with those of FEP and active inference, and discussed the relationship between them.
We also introduced state-of-the-art and research gaps of studies on world models and predictive in robotics. We described six frontiers and challenges, i.e., latent representations for action planning, neuro-symbolic predictive models, affordance perception, social interaction, brain-inspired world models, and cognitive architecture.
Through the survey and clarification of challenges, we provided future directions for developing cognitive and developmental robots based on world models and predictive coding.

\section*{Acknowledgements}
This work was partially supported by JST Moonshot R\&D, Grant Number JPMJMS2033 and JPMJMS2011, JST PRESTO Grant Number JPMJPR22C9, the BAGEP Award of the Science Academy, TUBITAK ARDEB 1001 program (project number: 120E274), the European Union’s Horizon 2020 Framework Programme for Research and Innovation under Specific Grant Agreements No. 945539 (Human Brain Project SGA3), No. 952215 (TAILOR), and No. 824153 (POTION), the European Research Council under the Grant Agreement No. 820213 (ThinkAhead), the Human Brain Project Specific Grant Agreement 3 grant (ID 643945539, for the SPIKEFERENCE project),  Deepself project under the Priority Programme “The Active Self” (SPP 2134), the project `COURAGE - A social media companion safeguarding and educating students' (No. 95563 and No. 9B145), and the Volkswagen Foundation inside the initiative Artificial Intelligence and the Society of the Future.

\bibliographystyle{tADR}
\bibliography{integrated}

\end{document}